\documentclass{article}

\usepackage{comment}
\usepackage[normalem]{ulem}
\usepackage[preprint]{aaai24}

\usepackage[utf8]{inputenc} 
\usepackage[T1]{fontenc}    
\usepackage{hyperref}       
\usepackage{url}            
\usepackage{booktabs}       
\usepackage[table,xcdraw]{xcolor}
\usepackage{amsfonts}       
\usepackage{nicefrac}       
\usepackage{microtype}      
\usepackage{xcolor}         
\usepackage[pdftex]{graphicx}
\usepackage{subcaption}
\usepackage{floatrow}
\usepackage{amsmath}
\usepackage{amssymb}
\usepackage{mathabx,bm}
\usepackage{changepage}

\usepackage{algorithm}
\usepackage{algorithmic}
\usepackage{booktabs}       

\title{Towards an On-device Agent for Text Rewriting}
\author{
Yun Zhu\thanks{Equal Contribution} \qquad
Yinxiao Liu \footnotemark[1] \qquad
Felix Stahlberg\qquad
Shankar Kumar\qquad
Yu-hui Chen\qquad \AND
Liangchen Luo\qquad
Lei Shu\qquad
Renjie Liu\qquad
Jindong Chen\qquad
Lei Meng \thanks{Correspondence to \{yunzhu, leimeng\}@google.com .}\\
Google Research
}

\usepackage{bibentry}

\begin{document}

\maketitle

\begin{abstract}

Large Language Models (LLMs) have demonstrated impressive capabilities for text rewriting.
Nonetheless, the large sizes of these models make them impractical for on-device inference, which would otherwise allow for enhanced privacy and economical inference.
Creating a smaller yet potent language model for text rewriting presents a formidable challenge because it requires balancing the need for a small size with the need to retain the emergent capabilities of the LLM, that requires costly data collection. To address the above challenge, we introduce a new instruction tuning approach for building a mobile-centric text rewriting model.
Our strategies enable the generation of high quality training data without any human labeling.
In addition, we propose a heuristic reinforcement learning framework which substantially enhances performance without requiring preference data. To further bridge the performance gap with the larger server-side model, we propose an effective approach that combines the mobile rewrite agent with the server model using a cascade.
To tailor the text rewriting tasks to mobile scenarios, we introduce \textsc{MessageRewriteEval}, a benchmark that focuses on text rewriting for messages through natural language instructions. 
Through empirical experiments, we demonstrate that our on-device model surpasses the current state-of-the-art LLMs in text rewriting while maintaining a significantly reduced model size. 
Notably, we show that our proposed cascading approach improves model performance.

\end{abstract}

\section{Introduction}
\label{sec:intro}
Text rewriting can be thought of as a variant of controlled text generation~\citep{zhang2022survey}, where textual inputs are altered based on the user's requirement. Various text rewriting categories have been extensively explored, including paraphrasing~\citep{siddique2020unsupervised, xu2012paraphrasing}, style transfer~\citep{riley2020textsettr, zhang2020parallel, reif2021recipe}, and sentence fusion~\citep{mallinson2022edit5}.
The advent of Large Language Models (LLMs)~\citep{palm2, brown2020language-gpt3, touvron2023llama} has ushered in a new era for text rewriting, demonstrating unparalleled quality by harnessing pre-trained models~\citep{shu2023rewritelm}. 
With the widespread use of mobile communications and text messaging~\citep{hanson2010cell, pennington2022toll},
these LLMs are being integrated into text rewriting applications, enabling users to create messages that are ``formal'', ``concise'' etc.~\citep{magiccompose}.


Despite the impressive text rewriting ability enabled by 
LLMs, their deployment for real-world chat messaging faces practical issues. The model size is often too large for on-device inference, yet using a server-based architecture presents challenges in maintaining users' privacy~\citep{li2021survey}, compromises offline operational capability~\citep{murshed2021machine}, and leads to higher overall compute costs~\citep{chen2023frugalgpt}. 
Developing a compact yet potent language model presents an array of challenges, arising from the  trade-offs between the requirement for a small size and a need for retaining the emergent capability of the LLM~\citep{wei2022emergent}, which requires costly data acquisition~\citep{kang2023distill}.

In this paper, we present a systematic approach for enhancing the rewriting capability of LLMs while adhering to size constraints to ensure reasonable on-device inference speeds.
We introduce a benchmark called \textsc{MessageRewriteEval}, compiled from human-donated message texts and rewrites with diverse language instructions. Unlike existing benchmarks for text rewriting such as \textsc{EditEval}~\citep{dwivedi-edit-2022} or \textsc{OpenRewriteEval}~\citep{shu2023rewritelm} which are derived from text sourced from paragraphs or long passages, our benchmark is designed to better represent daily conversational exchanges between individuals.  


Inspired by InstructGPT \citep{ouyang2022training}, we train our model using a combination of supervised fine-tuning (SFT) and reinforcement learning (RL). While InstructGPT relies heavily on human raters for both instruction data (for supervised fine-tuning) and preference data (for reward training), our approach minimizes human intervention in the data collection process. To elaborate: 
(1) For instruction data generation, we leverage hallucinations from LLMs to generate high quality synthetic data.  
(2) Instead of using synthetic data for training a reward model~\citep{shu2023rewritelm}, we propose a heuristic based reward signal for reinforcement learning that can improve the model without additional labeling. 
We conduct empirical investigations to assess the model's performance against the \textsc{MessageRewriteEval} benchmark. Our proposed model outperforms its corresponding foundation model by a substantial margin. Furthermore, it outperforms other instruction-tuned LLMs, which validates the usefulness of the generated training data and the proposed heuristic reinforcement learning.


To further close the gap between the server-side giant LLMs and their smaller on-device counterparts, we propose a cascading approach to chain our on-device model with the more powerful server model.  The system follows a simple yet effective principle: the server side will only be used when the on-device language model fails to provide a good response. Instead of relying on an external model to judge the quality of response~\citep{chen2023frugalgpt}, we propose to add a simple suffix to the on-device model output that indicates how confident the model is in its prediction. The suffix is learned from the larger server-side LLM via distillation. Our findings demonstrate that the proposed cascading approach further enhances performance.


Our main contributions can be summarized as follows:
\begin{itemize}
    \item We develop a powerful LLM that demonstrates superior performance compared to the state-of-the-art LLMs for text rewriting, while being efficient for on-device inference. Importantly, this model’s efficacy does not rely on human-labeled data collection. 
    We devise innovative strategies to generate varied instruction datasets for rewriting, that enhances the editing and rewriting capacities of the model. Additionally, we present a heuristic-based reinforcement learning approach that eliminates the need for training the reward model.
    \item We design an effective cascading mechanism to connect our on device model to the server side model. We distill the critiquing ability of the server LLM to the smaller model using discriminative training, which enables efficient inference. Our cascading strategy can further improve the on-device model's performance, bringing it closer to the capabilities of the server-side model while reducing the number of server calls.
    \item We introduce a new benchmark, \textsc{MessageRewriteEval}, designed for research on message text rewriting and covering different types of rewrites expressed through natural language instructions: formality, elaboration, shortening, paraphrasing, and proofreading. To the best of our knowledge, no such benchmark is currently available.


\end{itemize}

\section{Related Work}
\label{sec:related}

\subsection{Text Editing}
The text editing~\citep{51869} task covers a wide range of sub-tasks such as paraphrasing~\citep{huggingface:dataset:stsb_multi_mt}, style transfer~\citep{tikhonov2019style},  spelling and grammatical error correction~\citep{napoles2017gec}, formalization~\citep{rao2018dear}, simplification~\citep{xu2016optimizingsari} and elaboration~\citep{iv2022fruit}. 
Recent work has investigated a more diverse set of rewrite options~\cite{faltings2020text, schick2022peer, shu2023rewritelm} by leveraging the diversity of edits in Wikipedia.
Our model can also take diverse prompts as input but focuses on rewriting messages in a variety of ways, including formalizing, shortening, elaborating, paraphrasing, and proofreading. 
 

\subsection{Instruction Tuning}
Instruction tuning has been shown to improve model performance and generalization to unseen tasks~\citep{chung2022scaling,sanh2022multitask}.
InstructGPT~\citep{ouyang2022training} extends instruction tuning using reinforcement learning with human feedback (RLHF), which heavily relies on human raters to obtain instruction data and rankings of model outputs.
The dependency on human preference data could be alleviated by reinforcement learning with AI feedback (RLAIF)~\citep{bai2022constitutional, shu2023rewritelm}, but training a separate reward model is still required.  
We extend this framework using a heuristic based reinforcement learning~\citep{cheng2021heuristic} for rewriting tasks, which enables reinforcement learning without a reward model. 

\subsection{Distillation and Data Augmentation}
Knowledge distillation~\cite{hinton2015distilling} has been successfully used to transfer knowledge from more competent teacher models into smaller student models~\citep{hinton2015distilling,tang2019distilling,wang2021want,smith2022language,beyer2022knowledge,peng2023instruction, wu2023lamini}.
The quality of distillation could be improved in a variety of ways such as using a better design of Chain-of-Thought prompts~\citep{shu2023rewritelm}, combining the noisy predictions with majority vote~\citep{arora2022ask}, using a augmented label with reasoning~\citep{hsieh2023distilling}, reweighting the student’s loss~\cite{iliopoulos2022weighted} etc. 
Unlike previous work, we use a pre-trained LLM for generating reasonable paired data using model's ``hallucinations'' and then providing critique to the generated data for filtering. 
Furthermore, we extend our distillation technique to perform critiques.


\subsection{LLM Cascades}
Language model cascades have been investigated in many previous works~\citep{li2020cascadebert, cai2023large, wu2022ai, dohan2022language}. 
Frugal GPT~\citep{chen2023frugalgpt} proposed several strategies for using multiple LLMs to minimize the inference cost. For the cascaded design, the regression score from DistillBert~\citep{sanh2019distilbert} is used for deciding whether or not the model response is adequate. 
Although our approach achieves a similar goal, it does not require an extra model. We incorporate this capability into the language model in a single pass text generation step by using the suffixes of the generation~\citep{thoppilan2022lamda}. 

\section{Methods}
\label{sec:method}


Our approach follows the ``supervised fine-tuning (SFT) + reinforcement learning (RL)'' paradigm~\citep{ouyang2022training}, but does not require any human labeling or preference data collection. We first discuss our synthetic training data collection and supervised fine-tuning. Then we present our heuristic reward and RL process which further improves model performance. Lastly, we describe our cascaded method.


\subsection{Supervised Fine-tuning}
\label{subsection:training_data}
We follow existing works to leverage the document level edit data from Wikipedia~\citep{schick2022peer, shu2023rewritelm}. In pilot studies, we observed that using this data alone cannot provide adequate short form, message like data for training our on-device models. 
To generate in-domain data efficiently, we propose a data generation approach based on ``hallucinations'' of off-the-shelf LLMs, which can then be further filtered using the LLMs. The details of the training data are provided in the Appendix Section~\nameref{sec:appendix_traindata}.


\subsubsection{Paired Dataset from Hallucination}
\label{sec:hallucination}

\begin{figure}[t]
\includegraphics[width=0.9\linewidth]{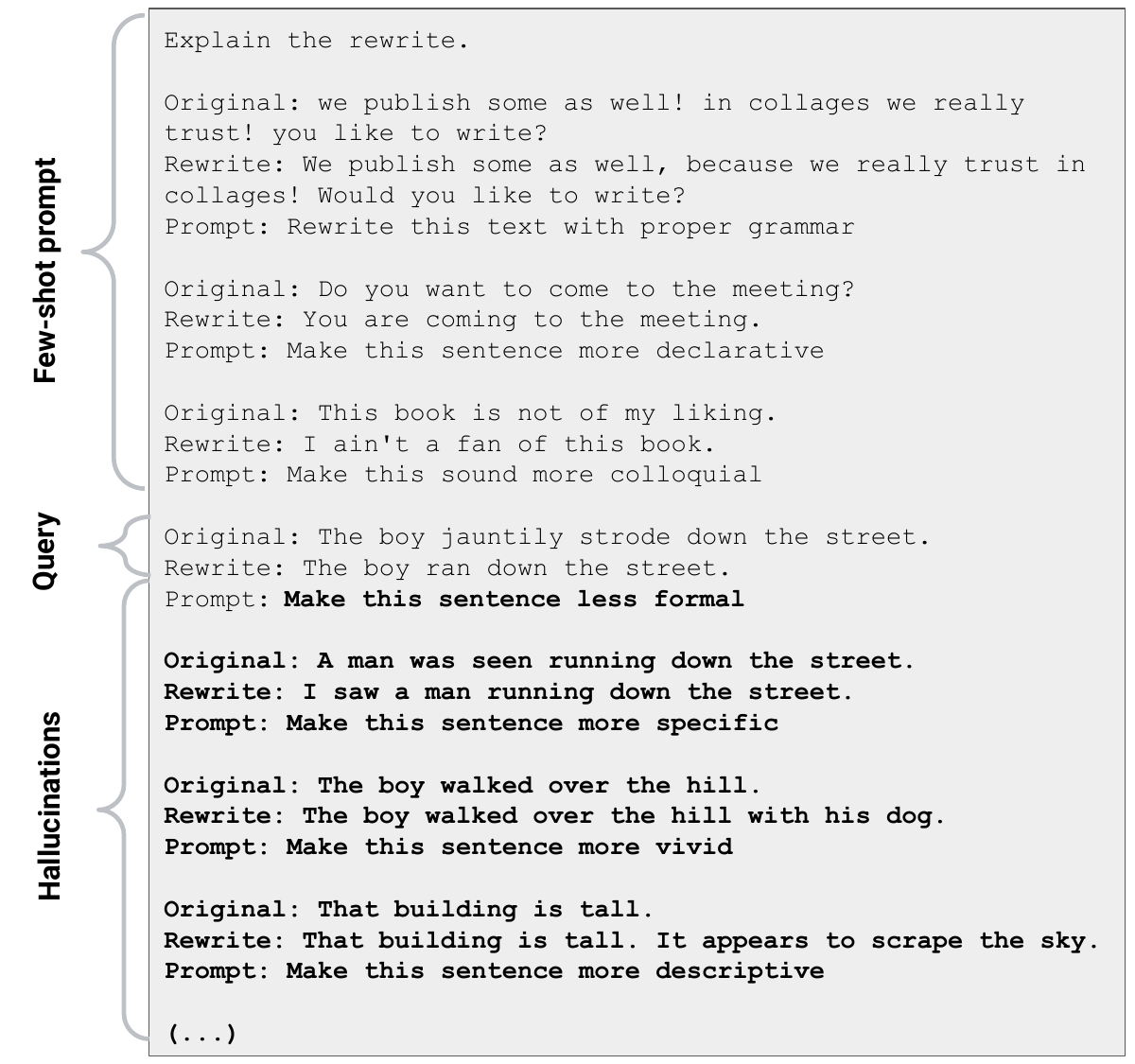}
\centering
\caption{Paired dataset generated from Hallucinations of the LLM. Bolded text includes a generated prompt for the query and the ``hallucinations'' of LLM, which contains hallucinated samples of Source, Rewrite, and Prompt.}
\label{fig:fs}
\end{figure}

To collect more shorter-form and message-like data, we leverage the few-shot capability of pre-trained LLMs.
Figure~\ref{fig:fs} shows an example of the initial prompts and demonstrates how the LLM is continuing to ``hallucinate'' diverse examples from a given query, which is sampled from a small seed query set. Hallucination is helping produce paired data in an efficient way. 



\subsubsection{An LLM guided data selection}
\label{data:selection}
To further improve the quality of our synthetic hallucination-based dataset, we propose to use LLMs to critique the generated data. 
We leverage the few-shot Chain-of-Thoughts (CoT) reasoning of the off-the-shelf LLM to judge whether the response is following the instruction of the prompt to rewrite the original sentence in a good manner. 
We provide detailed prompt samples in the Appendix in Table \ref{tab:critique_prompt}. 
We also leverage the self-consistency~\citep{wang2022self} approach to improve the accuracy of filtering. In practice, we only keep the data when it gets approval from all LLM judges.

\subsubsection{Generative Fine-tuning}
Given a pre-trained decoder-only language model, we fine-tune it using the collected instruction tuning dataset.
The input is formed by concatenating the \texttt{<instruction>} and the \texttt{<source>} with a newline, while the output is the \texttt{<target>}.

\subsection{Heuristic based Reinforcement Learning}
\label{subsection:ft}


The reinforcement learning part is typically called Reinforcement Learning with Human Feedback (RLHF)~\citep{rlhf} as human labelers are heavily involved in training the reward model.
In this section, we introduce a novel approach to improve alignment through heuristics without any human labeling.

\subsubsection{Heuristic Reward}
\label{heuristic_reward}
The intuition is that a few common heuristics can yield high quality rewrites.   
We propose to use the following heuristics as reward signals.
\begin{itemize}
    \item Natural Language Inference (\textbf{NLI})~\citep{bowman2015large} score over the source-prediction pair. Given a ``premise'' and a ``hypothesis'', NLI score
    is the probability that the ``hypothesis'' is correct given the ``premise''. In the context of LLMs, NLI score estimates whether or not the LLM's output prediction preserves meaning and factuality given the source text. We use the off-the-shelf NLI predictor introduced by~\citep{honovich2022true}. We denote the value as $nli$.
    \item \textbf{Reversed NLI}. Similar to NLI but the premise and the hypothesis are reversed. We denote it $rnli$.
    \item \textbf{Length Ratio}. The ratio of the number of tokens in the LLM output to that in the source text, which is denoted $length\_ratio$.
    \item \textbf{Edit Distance Ratio (Edit Ratio)}. Edit distance~\citep{levenshtein1966} measures the minimum number of token-level edits (insertions, deletions and substitutions) to convert a source text into a target text. We use the relative edit distance between the prediction and source text, we divide the edit distance by the length of the source text. The edit ratio represents the proportion of the source text that has been modified. We denote this as $edit\_ratio$.
    \item \textbf{N-gram frequency}. Text generation can easily get stuck in undesirable sentence-level loops with decoding~\citep{xu2022learning}. We propose measuring the N-gram frequency to detect potential loops in the generated output -- if the frequency of a certain N-gram is too high, we introduce a low reward to penalize it. The details are provided in Algorithm~\ref{alg:ngram}. We denote the output of this algorithm as $ngram\_reward$.
\end{itemize}

We formulate the final reward as a weighted combination of all the signals above in equation~(\ref{eq:reward}). For different rewriting tasks, the coefficient $\sigma_i$ should be designed to reflect the expectation of the rewrites. 
For instance, the expectation for ``shorten'' is higher $nli$ value (a larger positive $\sigma_1$)  and lower $length\_ratio$ (a negative $\sigma_3$). 
We share the choice of hyper-parameter $\sigma_i$  in Appendix Table~\ref{table:heu}.
\begin{equation}
\label{eq:reward}
\begin{split}
Reward = \sigma_1 nli + \sigma_2 rnli + \sigma_3 length\_ratio \\ 
+ \sigma_4 edit\_ratio + \sigma_5 ngram\_reward
\end{split}
\end{equation}


\begin{algorithm}[tb]
\caption{N-gram frequency based Reward}
\label{alg:ngram}
\begin{algorithmic}[1] 
\REQUIRE \textbf{Input}: $text$ \\
\REQUIRE \textbf{Input}: $NgramThresholdDict$\\
\ENSURE \textbf{Output}: $ngram\_reward$
\STATE words = text.split()
\FOR{$n, threshold \in NgramThresholdDict$}
\STATE $NgramHist = FindNgram(words, n)$
\STATE $MaxNgramCnt = max(NgramHist)$
\IF {$MaxNgramCnt >= threshold$} 
\STATE \textbf{return} $-C$
\ENDIF
\ENDFOR
\STATE \textbf{return} $0$
\end{algorithmic}
\end{algorithm}


\subsubsection{Reinforcement Learning}
We further refine the fine-tuned model by employing reinforcement learning~\citep{ouyang2022training}, guided by the heuristics provided.
The prompts for reinforcement learning are collected from the LLM hallucinations during training data generation. 
For each prompt in the train set, we first use LLM's fewshot ability to classify the prompt into the rewrite types. 
During the reinforcement learning, this rewrite type will be fed to the ``heuristic reward'' module to generate the reward, which will be finally optimized through PPO~\citep{schulman2017proximal}. 

\subsection{Critique Distillation and Model Cascade}
\label{subsection:ft1}

\begin{figure}[h]
\includegraphics[width=0.8\linewidth]{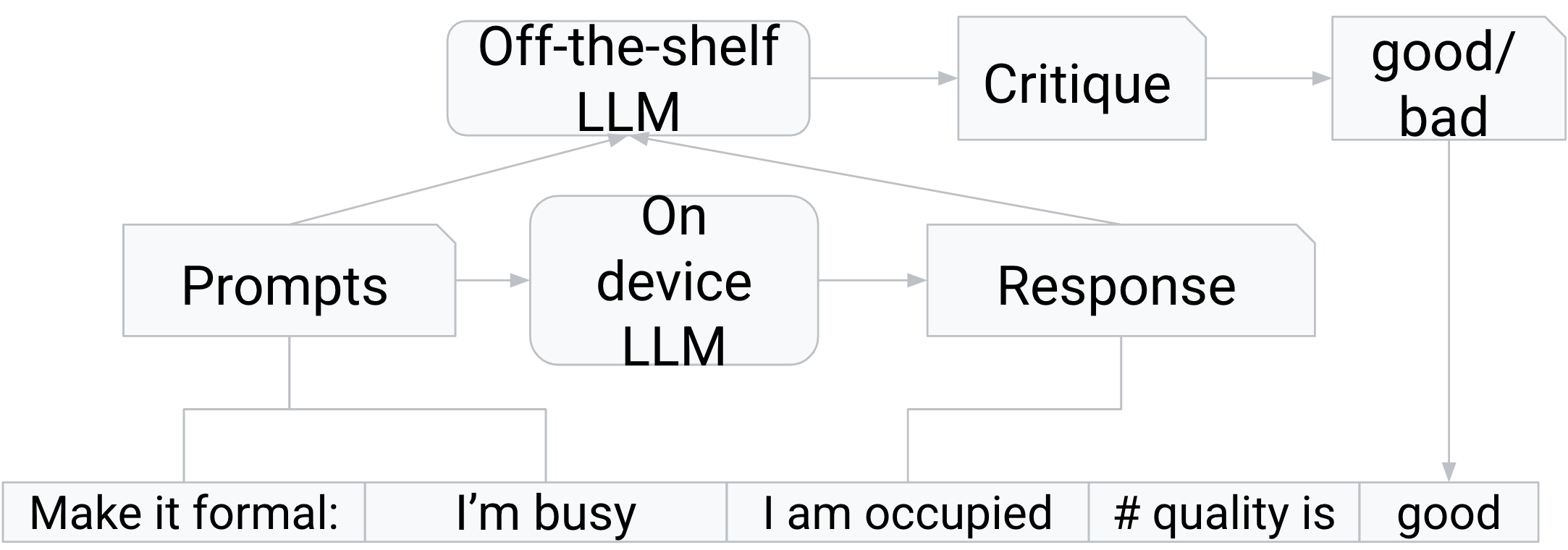}
\centering
\caption{The illustration of distillation for self-critiques. The final sentence with ``quality is good'' as suffix will be used as training data for discriminative training.}
\label{fig:suffix_distill}
\end{figure}

We apply a simple cascade mechanism whereby the on-device model serves as the first gate to the incoming rewrite request, and the large server side model is invoked only when the on-device rewrite is deemed low quality. Towards this goal, we need to answer two questions. First, how to enable the on-device model to do ``self-critique'', which is challenging given its small size and the complexity of the task.  Second, how do we make the process more efficient, without adding additional inference steps. 
We next present our suffix based distillation approach which provides a solution to the above questions.  
We leverage the off-the-shelf LLM as a critique and distill its knowledge as an extra ``suffix'' in the data. The approach is summarized in Figure~\ref{fig:suffix_distill}. 

\subsubsection{Critique Distillation from LLMs}

Similar to reinforcement learning, we prepare unpaired prompt data sampled from hallucinations of LLMs. The responses are generated by our on-device model. 
Then the (prompt, response) pairs are fed to the off-the-shelf LLM to decide whether they are acceptable or not. 
We leverage the Chain-of-Thought (CoT) reasoning along with the self-consistency approach. We use the prompts shown in Appendix Table \ref{tab:critique_prompt}

\subsubsection{Discriminative Fine-tuning}

Although generative fine-tuning with the larger LLM's response can make it possible to perform self-critiquing for the small models, ``generation'' and ``self-critique'' will be two separate text generation steps, resulting in increased inference times.  To fuse the two steps, we transform generative finetuning into discriminative fine-tuning~\citep{thoppilan2022lamda}.  
Suppose an example is classified as either ``good'' or ``bad'', this label is concatenated to the original prompt and response with some predefined delimiter. 
In this way, we can generate the suffix data that is distilled from the critique provided by the off-the-shelf LLM. 
Finally we finetune the ondevice model with the suffix data along with original generations. 

\subsubsection{Cascading}
Once the model is finetuned with suffix data, it can use the suffix score, i.e. probability of outputting ``good'', to decide whether to cascade. 
Specifically, after decoding some text we compare the ``suffix score'' $s$ (which is a probability between $0$ and $1$) and some pre-defined threshold $\gamma$. If $s > \gamma$, the on-device model is deemed confident; Otherwise, the model relies on the server side model.

\section{Experiment Settings}

\subsection{Model Training Setting}
Our pre-trained checkpoint is PaLM 2-XXS\footnote{We follow the size notations in PaLM 2 tech report~\citep{palm2}. Model size \textbf{XXS} is over 20 times smaller than model size \textbf{S} and over 5 times smaller than model size \textbf{XS}.}. We leverage pre-trained PALM 2-L as the off-the-shelf LLM for data generation, LLM filtering, and critique distillation.
The training hyper-parameters for instruction tuning and reinforcement learning are listed in the Appendix Section \nameref{sec:appendix_hyperparameter}. 




\subsection{Evaluation Datasets}

\subsubsection{MessageRewriteEval}
To evaluate the model performance in the on-device messaging scenario, we introduce \textsc{MessageRewriteEval}, a novel evaluation dataset specifically designed for message-level rewrite assessments. 
All text message pairs are sourced from real-life, human-written daily use cases and evaluated by human raters for data quality.
To ensure comprehensive evaluation, these pairs encompass five text rewrite tasks: \textit{Formalize}, \textit{Paraphrase}, \textit{Shorten}, \textit{Elaborate}, and \textit{Proofread}. Each text pair in the dataset consists of three components: \textit{source}, \textit{target}, and \textit{instruction}. The task distribution statistics and example instructions are provided in Appendix Section~\nameref{sec:message_rewrite_eval}. The data collection guidelines are given in Appendix Section~\nameref{sec:appendix_data_guide}.

\subsubsection{EditEval}
Besides the on-device messaging scenario, we evaluate the model performance on more general text rewriting tasks. We use the public rewrite benchmark EditEval \footnote{\url{https://github.com/facebookresearch/EditEval}} ~\citep{dwivedi-edit-2022} which covers rewriting task at both sentence and paragraph levels. The detailed description of the different datasets in this benchmark can be found in Appendix Section~\nameref{appendix:edit_eval}.

\subsection{Automatic Evaluation Metrics}
\label{subsec:automatic_metrics}

We employ various metrics to evaluate the model's quality:
\begin{itemize}
  \item \textbf{NLI}~\citep{bowman2015large} and \textbf{Reversed NLI} (Section~\nameref{heuristic_reward}).
  \item \textbf{Edit Distance Ratio (Edit Ratio)} (Section \nameref{heuristic_reward}).
  \item \textbf{SARI}~\citep{xu2016optimizingsari} is an n-gram based metric that measures the similarity of a prediction to both the source and reference texts. The scores of add, retain and delete operations are computed by averaging n-gram scores. The SARI metric is obtained using an arithmetic average of the F1 scores of add and retain operations and the precision of the delete operation.
\item \textbf{BLEU}~\citep{papineni2002bleu} is computed as a geometric mean of n-gram precisions of different orders. n-gram precision is the fraction of n-grams in the output that also occur in the reference. Though originally intended for machine translation, it is now widely used for evaluating rewrite tasks~\citep{xu2016simplication}. 
\item \textbf{Update-ROUGE (Updated-R)}~\citep{iv2022fruit} measures the recall of n-grams between the model's prediction and the references. It is a modified version of ROUGE~\citep{lin2003automaticrouge}. Updated-R specifically computes ROUGE-L on the updated sentences rather than the full text.
\item \textbf{Success Rate} We use the LLM to assess whether or not the response follows the instruction (i.e. ``good'' or not). Although a binary classification might be too coarse grained for evaluating rewrite quality, it is a very intuitive and straightforward metric to show the merit of cascading. The LLM prompts are provided in Appendix Table \ref{tab:critique_prompt}. For cascading experiments, we also measure \textbf{On-device Inference Ratio} which is the percentage of inference calls that are serviced by the on-device model. A higher ratio means a smaller percent of server calls. 
\end{itemize}




\subsection{Baselines}
\label{subsec:baselines_and_metrics}

Since it is designed for on-device application, our model has a compact size in comparison to other LLMs. In choosing baseline models, we prioritize the ones that are similar in size to ours.
We choose the state-of-the-art pre-trained models \textbf{PaLM 2}~\citep{palm2}, \textbf{LLaMA}~\citep{touvron2023llama} and the instruction tuned models \textbf{Alpaca}~\citep{alpaca}, \textbf{Vicuna}~\citep{vicuna2023}, \textbf{Flan-PaLM 2}~\citep{palm2} as our baseline models. We also provide \textbf{Alpaca-PaLM 2} for comparison. The Alpaca's instruction dataset is finetuned using a PaLM 2 baseline checkpoint.


For a fair comparison, we leveraged in-context learning with CoT few-shot prompting (we share the details in Appendix Section~\nameref{subsection:few_shot_prompts_inference}) to instruct the model to provide reasonable responses for the pre-trained models since they are not instruction tuned. In contrast, for the instruction tuned LLMs including ours, we use zero-shot settings.
For cascading, we note that constructing a powerful large language model is not within the the scope of this study. Therefore, our experiments utilize the 175B InsGPT~\citep{ouyang2022training} as the server model. 

\subsection{On-device Inference}
To demonstrate the effectiveness of running our models using limited resources, we obtain benchmark numbers on popular mobile phones to obtain two primary metrics: \textbf{Inference Latency per Token}, measured in milliseconds, and \textbf{Memory Consumption}, quantified in gigabytes during model operation.
We introduce an inference engine utilizing OpenCL that harnesses the computational capabilities of on-device GPUs. 
We adopt similar optimizations reported in~\cite{chen2023speed} and further devise special kernels tailored for our on-device Instruct-oriented models.
To accommodate models within constrained memory capacities, we employ 8-bit post-training quantization as the standard setting for reporting quality metrics. The latency/memory numbers of both 8- and lower-bit quantized model are presented to compare with commonly adopted configurations. We note that the quality implication of lower-bit quantization and quantization-aware training is beyond the scope of this paper. 


\section{Results}
\label{sec:exp}

\subsection{Results on EditEval}

Table~\ref{tab:perf_editeval} summarizes the results.
The metrics of the baseline models are directly obtained from the EditEval paper~\citep{dwivedi-edit-2022}. We list only those models whose sizes are similar to our on-device models; Nevertheless, our model is substantially smaller than these models. We provide SARI values for each dataset and extra Update-R scores for the two datasets relevant for the paragraph update task.

The results show that our on-device model with size \textbf{XXS} outperformed other models on most of the tasks despite being much smaller. 
For the fluency, coherence, paraphrase, simplification and paragraph update tasks, our model wins by a large margin. Heuristic reinforcement learning generally boosts the model's performance on all tasks.

\begin{adjustwidth}{-2cm}{} 
\begin{table}
\centering
\fontsize{7pt}{7pt}\selectfont
\setlength{\tabcolsep}{1pt}
\begin{tabular}{lcccccccccccccc}
\toprule
                              &  & JFL & $ITR_\text{FLU}$ & $ITR_\text{CLA}$ & $ITR_\text{COH}$ & STS  & TRK   & AST    & WNC   & \multicolumn{2}{c}{FRU} & \multicolumn{2}{c}{WFI} \\ \cmidrule{11-12} \cmidrule{13-14} 
                              &      & SARI     & SARI    & SARI  & SARI    & SARI    & SARI  & SARI  & SARI  & SARI   & Update-R    & SARI     & Updated-R    \\ \midrule
Copy                           & Size    & 26.7   & 32.3   & 29.5  & 31.3      & 21.1    & 26.3           & 20.7      & 31.9      & 29.8          & 0              & 33.6          & -              \\ \midrule
T0++~\citep{sanh2022multitask}                          & 11B  & 34.7    & 35.5     & 37.6  & 32.7     & 28.4       & 32.9           & 28.2       & 29.3       & 12.6          & 3.7            & 4.4           & 8.1            \\
PEER-11~\citep{schick2022peer}                        & 11B  & 55.8   & \textbf{52.1}    & 32.5  & 32.7    & 28.2     & 32.1           & 29.5   & \textbf{54.5}     & \textbf{39.6} & 31.4           & \textbf{34.9} & 20.4           \\ 
Tk~\citep{wang2022benchmarking-tk}                           & 3B   & 31.8     & 32.4   & \textbf{38.4}  & 33.8     & 30.2      & 32.8           & 29.9         & 31.1      & 12.6          & 3.6            & 1.3           & 4.5            \\
T0~\citep{sanh2022multitask}                            & 3B   & 42     & 24.6  & 32.6  & 22.2         & 34.3   & 34.4           & 32.3         & 22.3      & 14.2          & 9.6            & 5.1           & 16.3           \\

PEER-3~\citep{schick2022peer}                         & 3B   & 55.5   & 51.4    & 32.1  & 32.1     &28.6     & 32.5           & 30.5       & 53.3         & 39.1          & 30.9           & 34.4          & 18.7           \\
\midrule
PaLM 2~\citep{palm2}                    & S    & 36.07       & 22.68    & 28.79    & 27.82   & 34.45    & 34.32          & 35.92     & 25.2     & 24.28          & 26.39          & 11.41         & 20.42          \\
Flan PaLM 2~\citep{palm2}                    & XS    & 30.03       & 36.01    & 34.81    & 33.17   & 31.91    & 34.32          & 31.4    & 27.75    & 15.19   & 5.34   & 6.86   &  3.12          \\
Flan PaLM 2~\citep{palm2}                    & XXS    & 34.43       & 30.12    & 34.08    & 31.32   & 29.25    & 33.06          & 35.92    & 17.5    & 13.6               & 2.75          & 4.78         &  0.97          \\
Alpaca PaLM 2                   & XXS    & 29.33       & 17.01    & 24.42    & 23.81   & 32.59    & 31.56          & 33.46    & 28.11    & 23.53               & 14.22          & 6.5         &  3.72          \\
\midrule
SFT (Ours)                & XXS    & 58.36   & 37.67 & 33.85 & 36.03    & 37.49     & 38.88          & \textbf{41.95} & 32.35      & 35.44      & 47.49          & 22.03         & 32.36          \\
SFT + heuristic RL (Ours) & XXS    & \textbf{61.1}     & 40.26 & 34.81 & \textbf{37.33}      & \textbf{38.25}    & \textbf{40.21} & \textbf{41.95}          & 35.28     & 35.81      & \textbf{49.49} & 26.32         & \textbf{40.71} \\
\bottomrule
\end{tabular}
\caption{Model Performance on EditEval~\citep{dwivedi-edit-2022}. Only models with reasonable sizes are listed. Size \textbf{XXS} is less than half the size of T0/Tk models. Despite their reduced sizes, our models achieve even better performance than most of the other larger models. Relative to similar-sized instruction-tuned models, our models win by a large margin.}
\label{tab:perf_editeval}
\end{table}
\end{adjustwidth}

\subsection{Performance of the On-device Model}

To show that our approach can generally enhance the model's rewriting ability, we first report performance of our SFT model and RL model on \textsc{EditEval}. 
We then focus on message rewriting and evaluate the same SFT model and RL model on \textsc{MessageRewriteEval}.
Finally we present latency and memory metrics for on-device inference.

\subsection{Results on \textsc{MessageRewriteEval}}
\label{subsec:resultanalysis}

The automatic evaluation results for the \textsc{MessageRewriteEval} dataset are shown in Table \ref{tab:perf_auto_all}.  We first examine results of three sets of models: pre-trained LLMs, Instruction-Tuned LLMs and our on-device Instruction-Tuned LLMs.  

As discussed earlier, a larger Edit Ratio does not always correlate with better rewrite performance, as it may result from hallucinations. After conducting human inspection of the results, we observed that Edit Ratios larger than \textbf{0.2} often arise from undesired hallucinations. 
In terms of SARI, BLEU, and Update-R metrics, our on-device size models outperform LLaMA, Alpaca-7B and Vicuna-7B, despite having a much smaller size. We also compare our results to Alpaca-PaLM 2 and Flan-PaLM 2, which share the same base architecture and model size.
The fact that our model achieves much better SARI, BLEU, and Update-R scores validates the effectiveness of our approach. Moreover, the gap in performance between the SFT and RL models shows that our heurisic reinforment learning is very effective.

\begin{table*}[t]
\centering
\small
\setlength{\tabcolsep}{2.5pt}
\begin{tabular}{lccccccc}
\toprule
                                &  Size & Edit Ratio & NLI & Reversed NLI & SARI    & BLEU   & Update-R \\ \hline
                                 \multicolumn{8}{l}{\cellcolor[HTML]{DDDDDD}Giant LLM}      \\
 InsGPT~\citep{ouyang2022training}                & 175B                  & $0.18$     & $0.91$    & $0.88$   & $51.14$ & $35.0$ & $58.91$  \\
 \hline
\multicolumn{8}{l}{\cellcolor[HTML]{DDDDDD}Pre-trained LLMs}      \\
LLaMA~\citep{touvron2023llama}                     & 7B                  & $3.98$     & $0.68$    & $0.74$   & $31.58$ & $16.65$ & $29.24$   \\
PaLM 2~\citep{palm2}                       & XS                    & $0.98$     & $0.83$    & $0.72$  & $38.92$ & $22.98$ & $37.45$    \\
PaLM 2~\citep{palm2}                       & XXS                    & $1.59$     & $0.76$    & $0.82$  & $31.49$ & $18.81$ & $31.85$    \\ \hline
\multicolumn{8}{l}{\cellcolor[HTML]{DDDDDD}Instruction-Tuned LLMs}      \\
Alpaca~\citep{alpaca}                           & 7B                  & $0.26$     & $0.76$    & $0.76$  & $42.21$ & $24.80$ & $45.15$  \\
Vicuna~\citep{vicuna2023}                          & 7B                  & $1.27$     & $0.46$    & $0.52$   & $38.18$ & $14.30$ & $30.17$  \\ 
Flan-PaLM 2~\citep{palm2}                      & XS                  & $0.11$     & $\textbf{0.94}$    & $0.83$   & $29.50$ & $25.89$ & $34.63$   \\
Flan-PaLM 2~\citep{palm2}                      & XXS                  & $0.11$     & $0.93$    & $0.80$   & $27.41$ & $17.59$ & $15.43$   \\ 
Alpaca-PaLM 2                      & XXS                  & $0.3$     & $0.84$    & $0.78$   & $43.14$ & $25.88$ & $47.76$   \\ 
\hline
\multicolumn{8}{l}{\cellcolor[HTML]{DDDDDD}Our Intruction-Tuned On-device LLMs}      \\
SFT                & XXS                  & $\textbf{0.17}$     & $0.89$    & $0.75$   & $46.23$ & $27.78$ & $51.84$  \\
SFT  + heuristic RL                  & XXS                    & $0.16$     & $0.93$    & $\textbf{0.85}$   & $\textbf{47.14}$ & $\textbf{30.50}$ & $\textbf{54.97}$  \\ 
 \multicolumn{8}{l}{\cellcolor[HTML]{DDDDDD}Cascades}      \\
SFT  + heuristic RL   + critique distillation                 & XXS  & $0.16$     & $0.93$    & $0.84$   & $48.6$ & $32.43$ & $55.72$  \\
 Ours + InsGPT (40\% server calls)                & -                  & $0.16$     & $0.93$    & $\textbf{0.87}$   & $\textbf{49.87}$ & $\textbf{34.59}$ & $\textbf{58.87}$  \\
 Ours + InsGPT (15\% server calls)                & -                  & $0.16$     & $0.92$    & $0.86$   & $49.03$ & $33.76$ & $57.41$  \\
\bottomrule
\end{tabular}
\caption{Model Performance on \textsc{MessageRewriteEval}. Our models achieves best performance compared with all listed Pre-trained LLMs and Instruction-Tuned LLMs, which have either same or larger size then ours. When cascaded with InsGPT, the performance is further improved.}
\label{tab:perf_auto_all}
\end{table*}








\begin{figure*}[t]
    \begin{subfigure}{0.45\textwidth}
        \includegraphics[width=\textwidth]{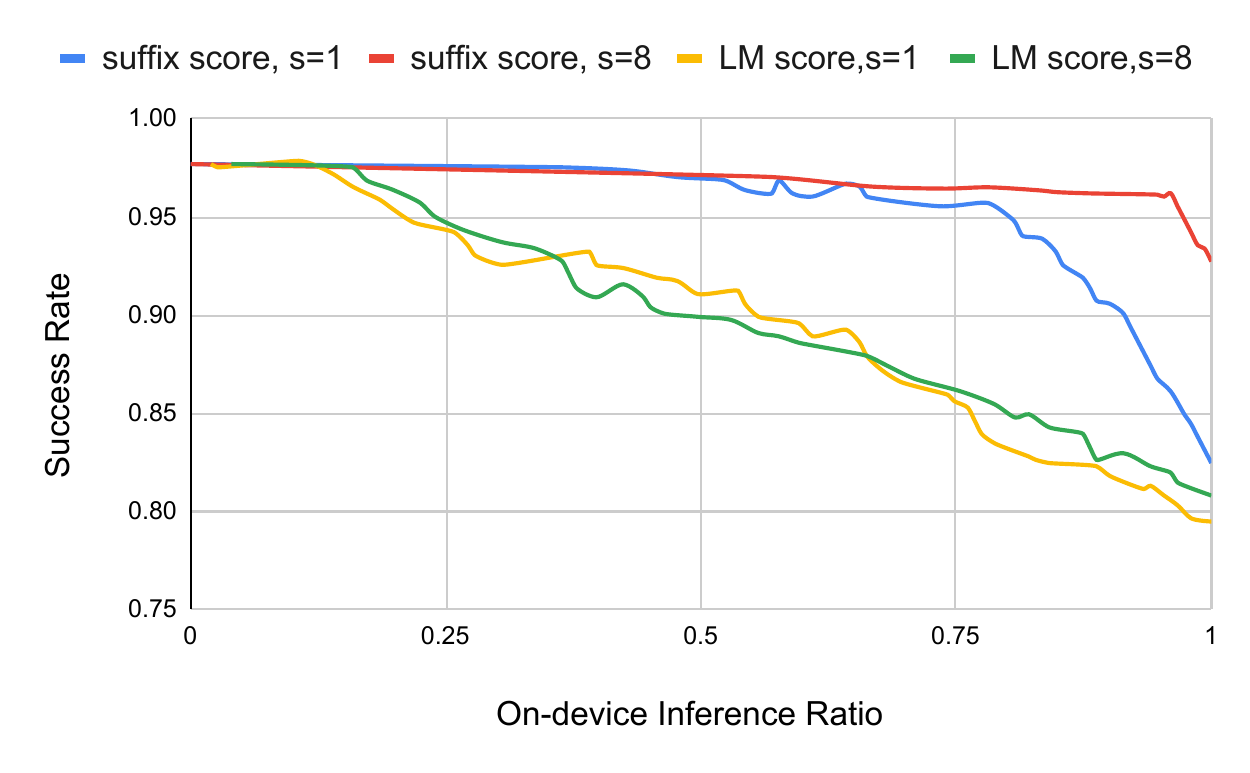}
        \caption{Suffix score VS LM score for paraphrase, with sample 1 or 8.}
        \label{fig:b}
    \end{subfigure}
    \hfill
    \begin{subfigure}{0.45\textwidth}
        \centering
        \includegraphics[width=\textwidth]{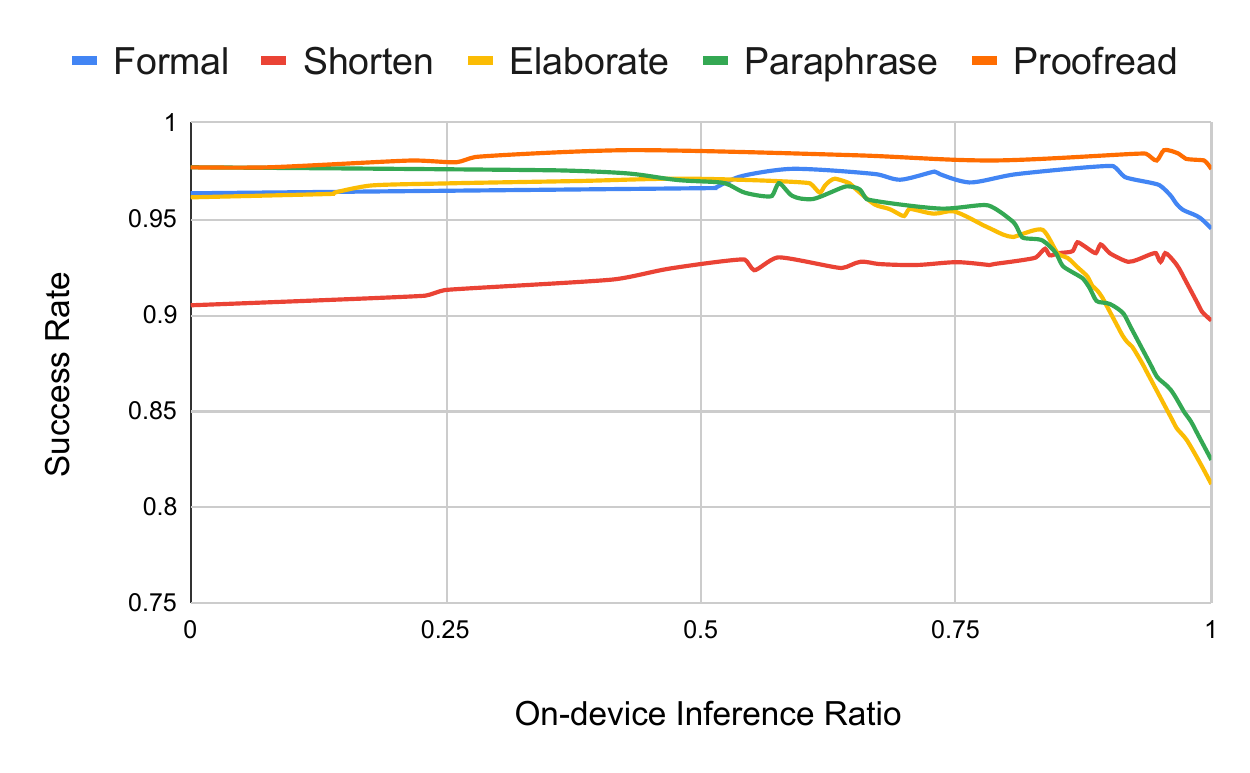}
     \caption{Evaluation of the cascades across different rewrite tasks. }
     \label{fig:a}
    \end{subfigure}
    \caption{Cascading our model with InsGPT on Rewriting Tasks. 
    Figure (a) demonstrates the benefit of using suffix score for cascading. Suffix score obtains better judgement for output quality than LM score when both decode with 1 sample; As a result, this capability could be further strengthened when decoding with multiple samples and choosing the one with highest score amongst the samples.
    Figure (b) evaluates the success rate with cascading across different rewrite types. The curves indicate that \textit{formalizing}, \textit{shortening} and \textit{proofreading} task obtain a smaller from cascading compared to \textit{paraphrasing} and \textit{elaborating}. With around 25\% server calls on the latter two tones, the success rate can be improved from 83\% to 95\%.  
}
    \label{fig:cascade}
\end{figure*}

\subsubsection{On-device Metrics}

\begin{table}[t]
\centering
\begin{tabular}{lcccc}
\toprule
& \multicolumn{2}{c}{S23} & \multicolumn{2}{c}{Pixel 7} \\
\cmidrule{2-3} \cmidrule{4-5}
& 8-bits & 4-bits & 8-bits & 4-bits \\ \midrule
P. Parsing (ms) & 1.2 & 1.2 & 4.2 & 4.2 \\
Decoding (ms) & 48.0 & 35.0 & 67.7 & 55.6 \\
Memory (Gb)  & 1.6 & 0.9 & 1.7 & 1.0 \\
\bottomrule
\end{tabular}
\caption{Benchmark results of our model. The average latency per token for the prompt parsing and decoding phases are reported in milliseconds. The last row shows the total memory consumption in gigabytes.}
\label{tab:benchmark}
\end{table}


Table~\ref{tab:benchmark} presents the performance benchmarks of our inference engine on both the Samsung S23 and Pixel 7 Pro. These evaluations were conducted using 1024 input tokens and decoding over 100 tokens. 
Results for both 8-bit and 4-bit quantized models are provided. It is noteworthy that, on the S23, the mean latency per token during the prompt parsing phase is 1.2ms (equivalent to >800 tokens/second), with the decoding latency being 35ms (29 tokens/second). To the best of our knowledge, the latency of our model on a cell phone is greatly faster than the reported numbers (i.e. 18 - 22 tokens/second) benchmarked on Macbook M1 Pro 32GB Ram for a 7B Llama model with 4-bits quantization~\citep{alpacaondevice}. 

\subsection{Performance of Cascading}
\label{result:cascade}

Our cascading experiments are conducted on the top of the on-device model with RL using \textsc{MessageRewriteEval} benchmark. Here we choose it over \textsc{EditEval} for cascading experiments as it is more aligned with the mobile cases.  
We first evaluate how the critique distillation is impacting the model's performance. Next we show the end-to-end cascading performance as a function of sampled thresholds $\gamma$.
We then present a detailed analysis and demonstrate that our suffix score is more effective than the baseline LM score. 
Finally we look into the performance for subcategories.

\subsubsection{The Effect of Critique Distillation}
In Table~\ref{tab:perf_auto_all} we show that the model's overall performance is further improved on SARI, BLEU, and Update-R with little regression on Reversed NLI when we combine the distilled discriminative dataset with the generative dataset. 
This suggests that with the suffix score from critique distillation, the model tends to pick sample with higher quality. 

\subsubsection{End-to-end Performance}
The on-device model's reliance on the server model is controlled by the threshold $\gamma$. As shown in Table~\ref{tab:perf_auto_all}, the performance of the cascaded models lies between the on-device and the server side model. With a higher number of server calls, we obtain higher SARI, BLEU, and Update-R, as expected. With 40\% server calls, the overall performance is already quite close to the full server model. 

\subsubsection{Suffix Score vs LM Score}
We now provide more insight into our cascading approach with suffix score.
We vary the threshold $\gamma$ from $0$ to $1$ to measure Success Rate as a function of the On-device Inference Ratio. 
The trade-off between the two metrics is shown in Figure~\ref{fig:cascade}.
To demonstrate the efficacy of the distilled suffix score derived from larger LLM critiques as a reliable indicator of output quality, we compare it with an LM score, representing the likelihood of the generated text.
As shown in Figure~\ref{fig:b}, ``suffix score with 1 sample'' is outperforming ``LM score with 1 sample'' by large margin. This indicates that given a text output,  suffix score offers higher quality estimates.
As a result, when sampling multiple outputs (8 samples), suffix score can accurately select the decoded candidate with the highest quality, which greatly improves performance. 
In contrast, the LM score stays almost unchanged when increasing the number of samples, showing that it is less helpful. 

\subsubsection{Variation across different Sub-categories}
We also compare the cascading effects for different rewriting sub-categories in Fig~\ref{fig:a}. 
The figure suggests that cascading may help paraphrase and elaborate tasks more than proofread, shorten and formalize. 
More importantly, around 25\% server calls can improve its success rate from 83\% to 95\%. 
\section{Conclusion}

In this paper we provided a systematic and effective approach to build an on-device rewrite agent.
We introduce \textsc{MessageRewriteEval}, a new benchmark that focuses on text rewriting for messages through natural language instructions.
We present a new instruction tuning approach for building a mobile-centric rewriting model, whose efficacy does not rely on human-labeled data or preference data.
We also develop an efficient and effective cascading approach using distillation of critiques.
Through experiments, we demonstrate that our on-device model outperforms the current state-of-the-art LLMs in text rewriting despite having a much smaller size. Furthermore,  cascading our model with the server side model can further boost its performance.


\section{Acknowledgments}
The authors would like to thank Abhanshu Sharma, Matt Sharifi, Victor Carbune, Lu Wang, Qifei Wang, Raman Sarokin, Yuanbo Zhang and Nevan Wichers for their insightful discussions and support. 

\bigskip


\bibliography{aaai24}

\begin{thebibliography}{66}
\expandafter\ifx\csname natexlab\endcsname\relax\def\natexlab#1{#1}\fi

\bibitem[{Arora et~al.(2022)Arora, Narayan, Chen, Orr, Guha, Bhatia, Chami,
  Sala, and R{\'e}}]{arora2022ask}
Simran Arora, Avanika Narayan, Mayee~F Chen, Laurel~J Orr, Neel Guha, Kush
  Bhatia, Ines Chami, Frederic Sala, and Christopher R{\'e}. 2022.
\newblock Ask me anything: A simple strategy for prompting language models.
\newblock \emph{arXiv preprint arXiv:2210.02441}.

\bibitem[{Bai et~al.(2022)Bai, Kadavath, Kundu, Askell, Kernion, Jones, Chen,
  Goldie, Mirhoseini, McKinnon et~al.}]{bai2022constitutional}
Yuntao Bai, Saurav Kadavath, Sandipan Kundu, Amanda Askell, Jackson Kernion,
  Andy Jones, Anna Chen, Anna Goldie, Azalia Mirhoseini, Cameron McKinnon,
  et~al. 2022.
\newblock Constitutional ai: Harmlessness from ai feedback.
\newblock \emph{arXiv preprint arXiv:2212.08073}.

\bibitem[{Beyer et~al.(2022)Beyer, Zhai, Royer, Markeeva, Anil, and
  Kolesnikov}]{beyer2022knowledge}
Lucas Beyer, Xiaohua Zhai, Am{\'e}lie Royer, Larisa Markeeva, Rohan Anil, and
  Alexander Kolesnikov. 2022.
\newblock Knowledge distillation: A good teacher is patient and consistent.
\newblock In \emph{Proceedings of the IEEE/CVF conference on computer vision
  and pattern recognition}, pages 10925--10934.

\bibitem[{Bowman et~al.(2015)Bowman, Angeli, Potts, and
  Manning}]{bowman2015large}
Samuel Bowman, Gabor Angeli, Christopher Potts, and Christopher~D Manning.
  2015.
\newblock A large annotated corpus for learning natural language inference.
\newblock In \emph{Proceedings of the 2015 Conference on Empirical Methods in
  Natural Language Processing}, pages 632--642.

\bibitem[{Brown et~al.(2020)Brown, Mann, Ryder, Subbiah, Kaplan, Dhariwal,
  Neelakantan, Shyam, Sastry, Askell et~al.}]{brown2020language-gpt3}
Tom Brown, Benjamin Mann, Nick Ryder, Melanie Subbiah, Jared~D Kaplan, Prafulla
  Dhariwal, Arvind Neelakantan, Pranav Shyam, Girish Sastry, Amanda Askell,
  et~al. 2020.
\newblock Language models are few-shot learners.
\newblock \emph{Advances in neural information processing systems},
  33:1877--1901.

\bibitem[{Burke(2023)}]{magiccompose}
Dave Burke. 2023.
\newblock Express yourself on android, with help from ai.
\newblock
  \url{https://blog.google/products/android/new-android-features-generative-ai/}.

\bibitem[{Cai et~al.(2023)Cai, Wang, Ma, Chen, and Zhou}]{cai2023large}
Tianle Cai, Xuezhi Wang, Tengyu Ma, Xinyun Chen, and Denny Zhou. 2023.
\newblock Large language models as tool makers.
\newblock \emph{arXiv preprint arXiv:2305.17126}.

\bibitem[{Chen et~al.(2023{\natexlab{a}})Chen, Zaharia, and
  Zou}]{chen2023frugalgpt}
Lingjiao Chen, Matei Zaharia, and James Zou. 2023{\natexlab{a}}.
\newblock Frugalgpt: How to use large language models while reducing cost and
  improving performance.
\newblock \emph{arXiv preprint arXiv:2305.05176}.

\bibitem[{Chen et~al.(2023{\natexlab{b}})Chen, Sarokin, Lee, Tang, Chang,
  Kulik, and Grundmann}]{chen2023speed}
Yu-Hui Chen, Raman Sarokin, Juhyun Lee, Jiuqiang Tang, Chuo-Ling Chang, Andrei
  Kulik, and Matthias Grundmann. 2023{\natexlab{b}}.
\newblock \href {http://arxiv.org/abs/2304.11267} {Speed is all you need:
  On-device acceleration of large diffusion models via gpu-aware
  optimizations}.

\bibitem[{Cheng et~al.(2021)Cheng, Kolobov, and
  Swaminathan}]{cheng2021heuristic}
Ching-An Cheng, Andrey Kolobov, and Adith Swaminathan. 2021.
\newblock Heuristic-guided reinforcement learning.
\newblock \emph{Advances in Neural Information Processing Systems},
  34:13550--13563.

\bibitem[{Chiang et~al.(2023)Chiang, Li, Lin, Sheng, Wu, Zhang, Zheng, Zhuang,
  Zhuang, Gonzalez, Stoica, and Xing}]{vicuna2023}
Wei-Lin Chiang, Zhuohan Li, Zi~Lin, Ying Sheng, Zhanghao Wu, Hao Zhang, Lianmin
  Zheng, Siyuan Zhuang, Yonghao Zhuang, Joseph~E. Gonzalez, Ion Stoica, and
  Eric~P. Xing. 2023.
\newblock \href {https://lmsys.org/blog/2023-03-30-vicuna/} {Vicuna: An
  open-source chatbot impressing gpt-4 with 90\%* chatgpt quality}.

\bibitem[{Chuklin et~al.(2022)Chuklin, Severyn, Mirylenka, Malmi, Stahlberg,
  Adamek, Mallinson, Krause, Kumar, and Dong}]{51869}
Aleksandr Chuklin, Aliaksei Severyn, Daniil Mirylenka, Eric~Emil Malmi, Felix
  Stahlberg, Jakub Adamek, Jonathan~Stephen Mallinson, Sebastian Krause,
  Shankar Kumar, and Yue Dong. 2022.
\newblock Text generation with text-editing models.
\newblock In \emph{Proceedings of NAACL 2022}.

\bibitem[{Chung et~al.(2022)Chung, Hou, Longpre, Zoph, Tay, Fedus, Li, Wang,
  Dehghani, Brahma et~al.}]{chung2022scaling}
Hyung~Won Chung, Le~Hou, Shayne Longpre, Barret Zoph, Yi~Tay, William Fedus,
  Eric Li, Xuezhi Wang, Mostafa Dehghani, Siddhartha Brahma, et~al. 2022.
\newblock Scaling instruction-finetuned language models.
\newblock \emph{arXiv preprint arXiv:2210.11416}.

\bibitem[{Dohan et~al.(2022)Dohan, Xu, Lewkowycz, Austin, Bieber, Lopes, Wu,
  Michalewski, Saurous, Sohl-Dickstein et~al.}]{dohan2022language}
David Dohan, Winnie Xu, Aitor Lewkowycz, Jacob Austin, David Bieber,
  Raphael~Gontijo Lopes, Yuhuai Wu, Henryk Michalewski, Rif~A Saurous, Jascha
  Sohl-Dickstein, et~al. 2022.
\newblock Language model cascades.
\newblock \emph{arXiv preprint arXiv:2207.10342}.

\bibitem[{Dwivedi-Yu et~al.(2022)Dwivedi-Yu, Schick, Jiang, Lomeli, Lewis,
  Izacard, Grave, Riedel, and Petroni}]{dwivedi-edit-2022}
Jane Dwivedi-Yu, Timo Schick, Zhengbao Jiang, Maria Lomeli, Patrick Lewis,
  Gautier Izacard, Edouard Grave, Sebastian Riedel, and Fabio Petroni. 2022.
\newblock \href {https://doi.org/10.48550/ARXIV.2209.13331} {Editeval: An
  instruction-based benchmark for text improvements}.
\newblock \emph{arXiv preprint arXiv:2305.01645}.

\bibitem[{Faltings et~al.(2020)Faltings, Galley, Hintz, Brockett, Quirk, Gao,
  and Dolan}]{faltings2020text}
Felix Faltings, Michel Galley, Gerold Hintz, Chris Brockett, Chris Quirk,
  Jianfeng Gao, and Bill Dolan. 2020.
\newblock Text editing by command.
\newblock \emph{arXiv preprint arXiv:2010.12826}.

\bibitem[{Gerganov(2023)}]{alpacaondevice}
Georgi Gerganov. 2023.
\newblock Port of facebook's llama model in c/c++.
\newblock \url{https://github.com/ggerganov/llama.cpp}.
\newblock Accessed: 2023-03-15.

\bibitem[{Hanson et~al.(2010)Hanson, Drumheller, Mallard, McKee, and
  Schlegel}]{hanson2010cell}
Trudy~L Hanson, Kristina Drumheller, Jessica Mallard, Connie McKee, and Paula
  Schlegel. 2010.
\newblock Cell phones, text messaging, and facebook: Competing time demands of
  today's college students.
\newblock \emph{College teaching}, 59(1):23--30.

\bibitem[{Hinton et~al.(2015)Hinton, Vinyals, and Dean}]{hinton2015distilling}
Geoffrey Hinton, Oriol Vinyals, and Jeff Dean. 2015.
\newblock Distilling the knowledge in a neural network.
\newblock \emph{arXiv preprint arXiv:1503.02531}.

\bibitem[{Honovich et~al.(2022)Honovich, Aharoni, Herzig, Taitelbaum,
  Kukliansy, Cohen, Scialom, Szpektor, Hassidim, and Matias}]{honovich2022true}
Or~Honovich, Roee Aharoni, Jonathan Herzig, Hagai Taitelbaum, Doron Kukliansy,
  Vered Cohen, Thomas Scialom, Idan Szpektor, Avinatan Hassidim, and Yossi
  Matias. 2022.
\newblock True: Re-evaluating factual consistency evaluation.
\newblock In \emph{Proceedings of the Second DialDoc Workshop on
  Document-grounded Dialogue and Conversational Question Answering}, pages
  161--175.

\bibitem[{Hsieh et~al.(2023)Hsieh, Li, Yeh, Nakhost, Fujii, Ratner, Krishna,
  Lee, and Pfister}]{hsieh2023distilling}
Cheng-Yu Hsieh, Chun-Liang Li, Chih-Kuan Yeh, Hootan Nakhost, Yasuhisa Fujii,
  Alexander Ratner, Ranjay Krishna, Chen-Yu Lee, and Tomas Pfister. 2023.
\newblock Distilling step-by-step! outperforming larger language models with
  less training data and smaller model sizes.
\newblock \emph{arXiv preprint arXiv:2305.02301}.

\bibitem[{Iliopoulos et~al.(2022)Iliopoulos, Kontonis, Baykal, Menghani, Trinh,
  and Vee}]{iliopoulos2022weighted}
Fotis Iliopoulos, Vasilis Kontonis, Cenk Baykal, Gaurav Menghani, Khoa Trinh,
  and Erik Vee. 2022.
\newblock Weighted distillation with unlabeled examples.
\newblock \emph{Advances in Neural Information Processing Systems},
  35:7024--7037.

\bibitem[{Iv et~al.(2022)Iv, Passos, Singh, and Chang}]{iv2022fruit}
Robert Iv, Alexandre Passos, Sameer Singh, and Ming-Wei Chang. 2022.
\newblock Fruit: Faithfully reflecting updated information in text.
\newblock In \emph{Proceedings of the 2022 Conference of the North American
  Chapter of the Association for Computational Linguistics: Human Language
  Technologies}, pages 3670--3686.

\bibitem[{Kang et~al.(2023)Kang, Xu, and Ritter}]{kang2023distill}
Junmo Kang, Wei Xu, and Alan Ritter. 2023.
\newblock Distill or annotate? cost-efficient fine-tuning of compact models.
\newblock \emph{arXiv preprint arXiv:2305.01645}.

\bibitem[{Levenshtein(1966)}]{levenshtein1966}
VI~Levenshtein. 1966.
\newblock {Binary Codes Capable of Correcting Deletions, Insertions and
  Reversals}.
\newblock \emph{Soviet Physics Doklady}, 10:707.

\bibitem[{Li et~al.(2020)Li, Lin, Chen, Ren, Li, Zhou, and
  Sun}]{li2020cascadebert}
Lei Li, Yankai Lin, Deli Chen, Shuhuai Ren, Peng Li, Jie Zhou, and Xu~Sun.
  2020.
\newblock Cascadebert: Accelerating inference of pre-trained language models
  via calibrated complete models cascade.
\newblock \emph{arXiv preprint arXiv:2012.14682}.

\bibitem[{Li et~al.(2021)Li, Wen, Wu, Hu, Wang, Li, Liu, and He}]{li2021survey}
Qinbin Li, Zeyi Wen, Zhaomin Wu, Sixu Hu, Naibo Wang, Yuan Li, Xu~Liu, and
  Bingsheng He. 2021.
\newblock A survey on federated learning systems: Vision, hype and reality for
  data privacy and protection.
\newblock \emph{IEEE Transactions on Knowledge and Data Engineering}.

\bibitem[{Lin and Hovy(2003)}]{lin2003automaticrouge}
Chin-Yew Lin and Eduard Hovy. 2003.
\newblock Automatic evaluation of summaries using n-gram co-occurrence
  statistics.
\newblock In \emph{Proceedings of the 2003 human language technology conference
  of the North American chapter of the association for computational
  linguistics}, pages 150--157.

\bibitem[{Mallinson et~al.(2022)Mallinson, Adamek, Malmi, and
  Severyn}]{mallinson2022edit5}
Jonathan Mallinson, Jakub Adamek, Eric Malmi, and Aliaksei Severyn. 2022.
\newblock Edit5: Semi-autoregressive text-editing with t5 warm-start.
\newblock \emph{arXiv preprint arXiv:2205.12209}.

\bibitem[{May(2021)}]{huggingface:dataset:stsb_multi_mt}
Philip May. 2021.
\newblock \href {https://github.com/PhilipMay/stsb-multi-mt} {Machine
  translated multilingual sts benchmark dataset.}

\bibitem[{Murshed et~al.(2021)Murshed, Murphy, Hou, Khan, Ananthanarayanan, and
  Hussain}]{murshed2021machine}
MG~Sarwar Murshed, Christopher Murphy, Daqing Hou, Nazar Khan, Ganesh
  Ananthanarayanan, and Faraz Hussain. 2021.
\newblock Machine learning at the network edge: A survey.
\newblock \emph{ACM Computing Surveys (CSUR)}, 54(8):1--37.

\bibitem[{Napoles et~al.(2017)Napoles, Sakaguchi, and
  Tetreault}]{napoles2017gec}
Courtney Napoles, Keisuke Sakaguchi, and Joel Tetreault. 2017.
\newblock \href {http://www.aclweb.org/anthology/E17-2037} {Jfleg: A fluency
  corpus and benchmark for grammatical error correction}.
\newblock In \emph{Proceedings of the 15th Conference of the European Chapter
  of the Association for Computational Linguistics: Volume 2, Short Papers},
  pages 229--234, Valencia, Spain. Association for Computational Linguistics.

\bibitem[{Ouyang et~al.(2022)Ouyang, Wu, Jiang, Almeida, Wainwright, Mishkin,
  Zhang, Agarwal, Slama, Ray et~al.}]{ouyang2022training}
Long Ouyang, Jeffrey Wu, Xu~Jiang, Diogo Almeida, Carroll Wainwright, Pamela
  Mishkin, Chong Zhang, Sandhini Agarwal, Katarina Slama, Alex Ray, et~al.
  2022.
\newblock Training language models to follow instructions with human feedback.
\newblock \emph{Advances in Neural Information Processing Systems},
  35:27730--27744.

\bibitem[{Papineni et~al.(2002)Papineni, Roukos, Ward, and
  Zhu}]{papineni2002bleu}
Kishore Papineni, Salim Roukos, Todd Ward, and Wei-Jing Zhu. 2002.
\newblock Bleu: a method for automatic evaluation of machine translation.
\newblock In \emph{Proceedings of the 40th annual meeting of the Association
  for Computational Linguistics}, pages 311--318.

\bibitem[{Passos et~al.(2023)Passos, Dai, Richter, Choquette, Sohn, So,
  Lepikhin, Taropa, Ni, Moreira, Mishra, Yu, Clark, Meier-Hellstern, Robinson,
  Vodrahalli, Omernick, Krikun, Moussalem, Johnson, Du, Firat, Bailey, Anil,
  Ruder, Shakeri, Qiao, Petrov, Garcia, Huang, Tay, Cheng, Wu, Xu, Zhang, and
  Nado}]{palm2}
Alex Passos, Andrew Dai, Bryan Richter, Christopher Choquette, Daniel Sohn,
  David So, Dmitry~(Dima) Lepikhin, Emanuel Taropa, Eric Ni, Erica Moreira,
  Gaurav Mishra, Jiahui Yu, Jon Clark, Kathy Meier-Hellstern, Kevin Robinson,
  Kiran Vodrahalli, Mark Omernick, Maxim Krikun, Maysam Moussalem, Melvin
  Johnson, Nan Du, Orhan Firat, Paige Bailey, Rohan Anil, Sebastian Ruder,
  Siamak Shakeri, Siyuan Qiao, Slav Petrov, Xavier Garcia, Yanping Huang,
  Yi~Tay, Yong Cheng, Yonghui Wu, Yuanzhong Xu, Yujing Zhang, and Zack Nado.
  2023.
\newblock Palm 2 technical report.
\newblock Technical report, Google Research.

\bibitem[{Peng et~al.(2023)Peng, Li, He, Galley, and Gao}]{peng2023instruction}
Baolin Peng, Chunyuan Li, Pengcheng He, Michel Galley, and Jianfeng Gao. 2023.
\newblock Instruction tuning with gpt-4.
\newblock \emph{arXiv preprint arXiv:2304.03277}.

\bibitem[{Pennington et~al.(2022)Pennington, Holmstrom, and
  Hall}]{pennington2022toll}
Natalie Pennington, Amanda~J Holmstrom, and Jeffrey~A Hall. 2022.
\newblock The toll of technology while working from home during covid-19.
\newblock \emph{Communication Reports}, 35(1):25--37.

\bibitem[{Rao and Tetreault(2018)}]{rao2018dear}
Sudha Rao and Joel Tetreault. 2018.
\newblock Dear sir or madam, may i introduce the gyafc dataset: Corpus,
  benchmarks and metrics for formality style transfer.
\newblock In \emph{Proceedings of the 2018 Conference of the North American
  Chapter of the Association for Computational Linguistics: Human Language
  Technologies, Volume 1 (Long Papers)}, pages 129--140.

\bibitem[{Reif et~al.(2021)Reif, Ippolito, Yuan, Coenen, Callison-Burch, and
  Wei}]{reif2021recipe}
Emily Reif, Daphne Ippolito, Ann Yuan, Andy Coenen, Chris Callison-Burch, and
  Jason Wei. 2021.
\newblock A recipe for arbitrary text style transfer with large language
  models.
\newblock \emph{arXiv preprint arXiv:2109.03910}.

\bibitem[{Riley et~al.(2020)Riley, Constant, Guo, Kumar, Uthus, and
  Parekh}]{riley2020textsettr}
Parker Riley, Noah Constant, Mandy Guo, Girish Kumar, David Uthus, and Zarana
  Parekh. 2020.
\newblock Textsettr: Few-shot text style extraction and tunable targeted
  restyling.
\newblock \emph{arXiv preprint arXiv:2010.03802}.

\bibitem[{Sanh et~al.(2019)Sanh, Debut, Chaumond, and
  Wolf}]{sanh2019distilbert}
Victor Sanh, Lysandre Debut, Julien Chaumond, and Thomas Wolf. 2019.
\newblock Distilbert, a distilled version of bert: smaller, faster, cheaper and
  lighter.
\newblock \emph{arXiv preprint arXiv:1910.01108}.

\bibitem[{Sanh et~al.(2022)Sanh, Webson, Raffel, Bach, Sutawika, Alyafeai,
  Chaffin, Stiegler, Le~Scao, Raja et~al.}]{sanh2022multitask}
Victor Sanh, Albert Webson, Colin Raffel, Stephen~H Bach, Lintang Sutawika,
  Zaid Alyafeai, Antoine Chaffin, Arnaud Stiegler, Teven Le~Scao, Arun Raja,
  et~al. 2022.
\newblock Multitask prompted training enables zero-shot task generalization.
\newblock In \emph{ICLR 2022-Tenth International Conference on Learning
  Representations}.

\bibitem[{Schick et~al.(2022)Schick, Dwivedi-Yu, Jiang, Petroni, Lewis,
  Izacard, You, Nalmpantis, Grave, and Riedel}]{schick2022peer}
Timo Schick, Jane Dwivedi-Yu, Zhengbao Jiang, Fabio Petroni, Patrick Lewis,
  Gautier Izacard, Qingfei You, Christoforos Nalmpantis, Edouard Grave, and
  Sebastian Riedel. 2022.
\newblock Peer: A collaborative language model.
\newblock \emph{arXiv preprint arXiv:2208.11663}.

\bibitem[{Schulman et~al.(2017)Schulman, Wolski, Dhariwal, Radford, and
  Klimov}]{schulman2017proximal}
John Schulman, Filip Wolski, Prafulla Dhariwal, Alec Radford, and Oleg Klimov.
  2017.
\newblock Proximal policy optimization algorithms.
\newblock \emph{arXiv preprint arXiv:1707.06347}.

\bibitem[{Shazeer and Stern(2018)}]{shazeer2018adafactor}
Noam Shazeer and Mitchell Stern. 2018.
\newblock Adafactor: Adaptive learning rates with sublinear memory cost.
\newblock In \emph{International Conference on Machine Learning}, pages
  4596--4604. PMLR.

\bibitem[{Shu et~al.(2023)Shu, Luo, Hoskere, Zhu, Liu, Tong, Chen, and
  Meng}]{shu2023rewritelm}
Lei Shu, Liangchen Luo, Jayakumar Hoskere, Yun Zhu, Canoee Liu, Simon Tong,
  Jindong Chen, and Lei Meng. 2023.
\newblock Rewritelm: An instruction-tuned large language model for text
  rewriting.
\newblock \emph{arXiv preprint arXiv:2305.15685}.

\bibitem[{Siddique et~al.(2020)Siddique, Oymak, and
  Hristidis}]{siddique2020unsupervised}
AB~Siddique, Samet Oymak, and Vagelis Hristidis. 2020.
\newblock Unsupervised paraphrasing via deep reinforcement learning.
\newblock In \emph{Proceedings of the 26th ACM SIGKDD international conference
  on knowledge discovery \& data mining}, pages 1800--1809.

\bibitem[{Smith et~al.(2022)Smith, Fries, Hancock, and
  Bach}]{smith2022language}
Ryan Smith, Jason~A Fries, Braden Hancock, and Stephen~H Bach. 2022.
\newblock Language models in the loop: Incorporating prompting into weak
  supervision.
\newblock \emph{arXiv preprint arXiv:2205.02318}.

\bibitem[{Tang et~al.(2019)Tang, Lu, Liu, Mou, Vechtomova, and
  Lin}]{tang2019distilling}
Raphael Tang, Yao Lu, Linqing Liu, Lili Mou, Olga Vechtomova, and Jimmy Lin.
  2019.
\newblock Distilling task-specific knowledge from bert into simple neural
  networks.
\newblock \emph{arXiv preprint arXiv:1903.12136}.

\bibitem[{Taori et~al.(2023)Taori, Gulrajani, Zhang, Dubois, Li, Guestrin,
  Liang, and Hashimoto}]{alpaca}
Rohan Taori, Ishaan Gulrajani, Tianyi Zhang, Yann Dubois, Xuechen Li, Carlos
  Guestrin, Percy Liang, and Tatsunori~B. Hashimoto. 2023.
\newblock Stanford alpaca: An instruction-following llama model.
\newblock \url{https://github.com/tatsu-lab/stanford_alpaca}.

\bibitem[{Thoppilan et~al.(2022)Thoppilan, De~Freitas, Hall, Shazeer,
  Kulshreshtha, Cheng, Jin, Bos, Baker, Du et~al.}]{thoppilan2022lamda}
Romal Thoppilan, Daniel De~Freitas, Jamie Hall, Noam Shazeer, Apoorv
  Kulshreshtha, Heng-Tze Cheng, Alicia Jin, Taylor Bos, Leslie Baker, Yu~Du,
  et~al. 2022.
\newblock Lamda: Language models for dialog applications.
\newblock \emph{arXiv preprint arXiv:2201.08239}.

\bibitem[{Tikhonov et~al.(2019)Tikhonov, Shibaev, Nagaev, Nugmanova, and
  Yamshchikov}]{tikhonov2019style}
Alexey Tikhonov, Viacheslav Shibaev, Aleksander Nagaev, Aigul Nugmanova, and
  Ivan~P Yamshchikov. 2019.
\newblock Style transfer for texts: Retrain, report errors, compare with
  rewrites.
\newblock In \emph{Proceedings of the 2019 Conference on Empirical Methods in
  Natural Language Processing and the 9th International Joint Conference on
  Natural Language Processing (EMNLP-IJCNLP)}, pages 3936--3945.

\bibitem[{Touvron et~al.(2023)Touvron, Lavril, Izacard, Martinet, Lachaux,
  Lacroix, Rozi{\`e}re, Goyal, Hambro, Azhar et~al.}]{touvron2023llama}
Hugo Touvron, Thibaut Lavril, Gautier Izacard, Xavier Martinet, Marie-Anne
  Lachaux, Timoth{\'e}e Lacroix, Baptiste Rozi{\`e}re, Naman Goyal, Eric
  Hambro, Faisal Azhar, et~al. 2023.
\newblock Llama: Open and efficient foundation language models.
\newblock \emph{arXiv preprint arXiv:2302.13971}.

\bibitem[{Wang et~al.(2021)Wang, Liu, Xu, Zhu, and Zeng}]{wang2021want}
Shuohang Wang, Yang Liu, Yichong Xu, Chenguang Zhu, and Michael Zeng. 2021.
\newblock Want to reduce labeling cost? gpt-3 can help.
\newblock \emph{arXiv preprint arXiv:2108.13487}.

\bibitem[{Wang et~al.(2022{\natexlab{a}})Wang, Wei, Schuurmans, Le, Chi,
  Narang, Chowdhery, and Zhou}]{wang2022self}
Xuezhi Wang, Jason Wei, Dale Schuurmans, Quoc Le, Ed~Chi, Sharan Narang,
  Aakanksha Chowdhery, and Denny Zhou. 2022{\natexlab{a}}.
\newblock Self-consistency improves chain of thought reasoning in language
  models.
\newblock \emph{arXiv preprint arXiv:2203.11171}.

\bibitem[{Wang et~al.(2022{\natexlab{b}})Wang, Mishra, Alipoormolabashi, Kordi,
  Mirzaei, Arunkumar, Ashok, Dhanasekaran, Naik, Stap
  et~al.}]{wang2022benchmarking-tk}
Yizhong Wang, Swaroop Mishra, Pegah Alipoormolabashi, Yeganeh Kordi, Amirreza
  Mirzaei, Anjana Arunkumar, Arjun Ashok, Arut~Selvan Dhanasekaran, Atharva
  Naik, David Stap, et~al. 2022{\natexlab{b}}.
\newblock Benchmarking generalization via in-context instructions on 1,600+
  language tasks.
\newblock \emph{arXiv preprint arXiv:2204.07705}.

\bibitem[{Wei et~al.(2022)Wei, Tay, Bommasani, Raffel, Zoph, Borgeaud,
  Yogatama, Bosma, Zhou, Metzler et~al.}]{wei2022emergent}
Jason Wei, Yi~Tay, Rishi Bommasani, Colin Raffel, Barret Zoph, Sebastian
  Borgeaud, Dani Yogatama, Maarten Bosma, Denny Zhou, Donald Metzler, et~al.
  2022.
\newblock Emergent abilities of large language models.
\newblock \emph{arXiv preprint arXiv:2206.07682}.

\bibitem[{Wei~Xu and Callison-Burch(2016)}]{xu2016simplication}
Ellie Pavlick Quanze~Chen Wei~Xu, Courtney~Napoles and Chris Callison-Burch.
  2016.
\newblock Optimizing statistical machine translation for text simplification.
\newblock \emph{Transactions of the Association for Computational Linguistics},
  4.

\bibitem[{Wu et~al.(2023)Wu, Waheed, Zhang, Abdul-Mageed, and
  Aji}]{wu2023lamini}
Minghao Wu, Abdul Waheed, Chiyu Zhang, Muhammad Abdul-Mageed, and Alham~Fikri
  Aji. 2023.
\newblock Lamini-lm: A diverse herd of distilled models from large-scale
  instructions.
\newblock \emph{arXiv preprint arXiv:2304.14402}.

\bibitem[{Wu et~al.(2022)Wu, Terry, and Cai}]{wu2022ai}
Tongshuang Wu, Michael Terry, and Carrie~Jun Cai. 2022.
\newblock Ai chains: Transparent and controllable human-ai interaction by
  chaining large language model prompts.
\newblock In \emph{Proceedings of the 2022 CHI conference on human factors in
  computing systems}, pages 1--22.

\bibitem[{Xu et~al.(2022)Xu, Liu, Yan, Cai, Li, and Li}]{xu2022learning}
Jin Xu, Xiaojiang Liu, Jianhao Yan, Deng Cai, Huayang Li, and Jian Li. 2022.
\newblock Learning to break the loop: Analyzing and mitigating repetitions for
  neural text generation.
\newblock \emph{Advances in Neural Information Processing Systems},
  35:3082--3095.

\bibitem[{Xu et~al.(2016)Xu, Napoles, Pavlick, Chen, and
  Callison-Burch}]{xu2016optimizingsari}
Wei Xu, Courtney Napoles, Ellie Pavlick, Quanze Chen, and Chris Callison-Burch.
  2016.
\newblock Optimizing statistical machine translation for text simplification.
\newblock \emph{Transactions of the Association for Computational Linguistics},
  4:401--415.

\bibitem[{Xu et~al.(2012)Xu, Ritter, Dolan, Grishman, and
  Cherry}]{xu2012paraphrasing}
Wei Xu, Alan Ritter, William~B Dolan, Ralph Grishman, and Colin Cherry. 2012.
\newblock Paraphrasing for style.
\newblock In \emph{Proceedings of COLING 2012}, pages 2899--2914.

\bibitem[{Zhang et~al.(2022)Zhang, Song, Li, Zhou, and Song}]{zhang2022survey}
Hanqing Zhang, Haolin Song, Shaoyu Li, Ming Zhou, and Dawei Song. 2022.
\newblock A survey of controllable text generation using transformer-based
  pre-trained language models.
\newblock \emph{arXiv preprint arXiv:2201.05337}.

\bibitem[{Zhang et~al.(2020)Zhang, Ge, and Sun}]{zhang2020parallel}
Yi~Zhang, Tao Ge, and Xu~Sun. 2020.
\newblock Parallel data augmentation for formality style transfer.
\newblock \emph{arXiv preprint arXiv:2005.07522}.

\bibitem[{Ziegler et~al.(2019)Ziegler, Stiennon, Wu, Brown, Radford, Amodei,
  Christiano, and Irving}]{rlhf}
Daniel~M. Ziegler, Nisan Stiennon, Jeffrey Wu, Tom~B. Brown, Alec Radford,
  Dario Amodei, Paul~F. Christiano, and Geoffrey Irving. 2019.
\newblock \href {http://arxiv.org/abs/1909.08593} {Fine-tuning language models
  from human preferences}.
\newblock \emph{CoRR}, abs/1909.08593.

\end{thebibliography}
\bibliographystyle{acl_natbib}

\clearpage
\appendix
\section{Appendix}
\label{sec:appendix}

\subsection{\textsc{MessageRewriteEval} Data}
\label{sec:message_rewrite_eval}

\textbf{Statistics} of the \textsc{MessageRewriteEval} are located in Table \ref{tab:data}. For every task and the complete dataset, we offer the following details: sample counts; the average word length for instruction (Ins), source (Sou), and target (Tar); the average length ratio (Len Ra) of the target over the source; and the Edit Ratio (Edit Ra, refer to Section \nameref{subsec:automatic_metrics}). All these statistical measurements are based on words. Additionally, NLI scores between the source and the golden target are available in both directions: from source to target and from target to source. Besides, samples of the instructions for each task in \textsc{MessageRewriteEval} are presented in Table \ref{tab:instruction_sample}. 

\begin{table}[h]
\centering
\fontsize{9pt}{9pt}\selectfont
\setlength{\tabcolsep}{1.2pt}
\begin{tabular}{lccccccccc}
\toprule
           &        &          &          &          &           &           &            \multicolumn{2}{c}{NLI} \\ \cmidrule{8-9} 
           & Size   & Ins & Sou   & Tar   & Len Ra & Edit Ra & s-t    & t-s    \\ \midrule
$\textit{Formalize}$  & $177$  & $5.42$	& $8.86$	& $12.3$	& $1.3$   & $0.26$	 &	$0.79$	 &	$0.83$     \\
$\textit{Shorten}$    & $221$  & $5.33$	& $9.65$	& $5.92$	& $0.6$    & $0.21$	 &	$0.9$	 &	$0.88$     \\
$\textit{Elaborate}$  & $206$  & $5.76$	& $9.42$	& $29.27$	& $3.1$    & $0.15$	 &	$0.95$	 &	$0.8$     \\
$\textit{Paraphrase}$ & $151$  & $3.83$	& $9.58$	& $10.83$	& $1.1$    & $0.21$	 &	$0.9$	 &	$0.88$     \\
$\textit{Proofread}$ & $280$  & $11.64$	& $10.88$	& $10.24$	& $0.94$    & $0.12$ &	$0.95$	 &	$0.96$     \\ \midrule
All        & $1035$ & $6.92$   & $9.79$ & $13.54$ & $1.38$   & $0.18$	 &	$0.91$	 &	$0.88$     \\ 
\bottomrule
\end{tabular}
\caption{
Statistics of \textsc{MessageRewriteEval}.
	}
\label{tab:data}
\end{table}

\begin{table}[h]
\centering
\fontsize{9pt}{9pt}\selectfont
\setlength{\tabcolsep}{1.2pt}
\begin{tabular}{p{0.15\linewidth}p{0.8\linewidth}}
\toprule
Task & \hspace{0.5cm}Instruction Examples\\ 
\midrule

$\textit{Formalize}$       &    \hspace{0.5cm}\begin{tabular}[c]{@{}l@{}}Make the text formal.\\
Make this sentence more formal.\\
Formalize the text.\\
Rewrite this sentence in a more formal way.
\end{tabular}\\\midrule
$\textit{Shorten}$       & \hspace{0.5cm}\begin{tabular}[c]{@{}l@{}}Make the text more concise.\\
Rewrite this text in concise language.\\
Make the text shorter.\\
Make this sound more concise\end{tabular}\\\midrule
$\textit{Elaborate}$      & \hspace{0.5cm}\begin{tabular}[c]{@{}l@{}}Make this more verbose.\\
Expand this text.\\
Rephrase this sentence in a more expand style.\\
Make the text more elaborated.\end{tabular}\\\midrule
$\textit{Paraphrase}$   &  \hspace{0.5cm}\begin{tabular}[c]{@{}l@{}}Rewrite this sentence.\\
Rephrase the text.\\
Paraphrase the following text.\\
Rewrite, reword and reorganize. way.
\end{tabular}\\\midrule
$\textit{Proofread}$  & \hspace{0.5cm}\begin{tabular}[c]{@{}p{0.8\linewidth}@{}}Fix the grammar error or spelling error of the following text.\\
Correct the following sentence if there is any spelling or grammar error.\\
Please proofread this sentence.
\end{tabular} \\

\bottomrule

\end{tabular}
\caption{
The instruction samples for each task of \textsc{MessageRewriteEval}.	}
\label{tab:instruction_sample}
\end{table}

\subsection{Data Guidelines}
\label{sec:appendix_data_guide}
During the data donation and review process for \textsc{MessageRewriteEval}, the follow guideline is provided:
\begin{itemize}
  \item Content should be preserved in target from source.
  \item For certain rewrite task, the target should follow the requirement in the instruction.
  \item \textit{Formalize}: the target should be more formal compared to source including: (1) formal vocabulary, (2) impersonal expression and (3) standard grammatical forms.
  \item \textit{Shorten}: the target is simpler, more concise compared to source preserving the tone and format from the source.
  \item \textit{Elaborate}: the target extend the source with more relevant information and ideas but the same tone and format as the source. The relevant information should not be made up facts.
  \item \textit{Paraphrase}: the target changes the wording of the source while preserving the content, tone, format and verbosity.
  \item \textit{Proofread}: the target fixes the grammar and wording errors in the source text.
\end{itemize}

\subsection{EditEval Dataset}
\label{appendix:edit_eval}
The rewrite task and dataset information in EditEval benchmark can be found in Table \ref{tab:editeval_data}. The two datasets for \textit{Updating} task are paragraph level, while the rest datasets are all sentence level.

\begin{table}[h]
\begin{tabular}{lccc}
\toprule
Task           & Dataset      & Abbrev. & Size \\ \midrule
\textit{Fluency}        & JFLEG        & JFL     & 747  \\
\textit{Fluency}        & ITERATOR     & $ITR_\textit{FLU}$  & 203 \\
\textit{Clarity}        & ITERATOR     & $ITR_\textit{CLA}$  & 342 \\
\textit{Coherence}      & ITERATOR     & $ITR_\textit{COH}$  & 76 \\
\textit{Simplification} & ASSET        & AST     & 359  \\
\textit{Simplification} & TurkCorpus   & TRK     & 359  \\
\textit{Paraphrasing}   & STS          & STS     & 97   \\
\textit{Neutralization} & WNC          & WNC     & 1000 \\
\textit{Updating}       & FRUIT        & FRU     & 914  \\
\textit{Updating}       & WAFER-INSERT & WFI     & 4565 \\
\bottomrule
\end{tabular}
\caption{EditEval Dataset Statistics}
\label{tab:editeval_data}
\end{table}

\subsection{Hyper-parameter Setting}
\label{sec:appendix_hyperparameter}
During supervised finetuning, SFT, we use $8$ Tensor Processing Units (TPU) V3 chips for fine-tuning. The batch size is $64$, and the maximum training step is $30000$. We use the Adafactor optimizer~\citep{shazeer2018adafactor} with a learning rate of $0.003$. Both the input and output sequence lengths are set to $1024$ tokens. 
The training dropout rate is $0.05$. 
For reinforcement learning, we compute the heuristic reward with parameters in ~\ref{table:heu}. We use same setup as fine-tuning except that the training step is $3000$. During inference, the temperature is set to $0.5$. Unless specifically noted, we use sampling decoding with sample number $8$ for our experiments.

\begin{table}[!t]
\centering
\begin{tabular}{l|c|c|c|c|c}
    \hline
    Rewrites & 
    \multicolumn{1}{|p{1.5cm}|}{\centering NLI $\sigma_1$} & 
    \multicolumn{1}{|p{1.5cm}|}{\centering Reverse NLI $\sigma_2$} & 
    \multicolumn{1}{|p{1.5cm}|}{\centering Length Ratio $\sigma_3$} & 
    \multicolumn{1}{|p{1.5cm}|}{\centering Edit Dist $\sigma_4$} & 
    \multicolumn{1}{|p{1.5cm}}{\centering Ngram Freq $\sigma_5$}  \\
    \hline
    Formalize & 1.0 & 1.0 & 0.0  & 0.4 & 1.0 \\
    Shorten & 1.0 & 0.4 & -0.2 & 0.4 & 1.0\\
    Elaborate & 0.4 & 1.0 & 0.5 & 0.4 & 1.0\\
    Paraphrase & 1.0 & 1.0 & 0.0 & 0.4 & 1.0\\
    proofread & 1.0 & 1.0 & 0.0 & 0.0 & 1.0\\
    \hline
\end{tabular}
\caption{The choice of $sigma_i$. For formalize and paraphrase, the length ratio is not considered important while for proofread/grammar correction, we apply the additional logic that the length ratio should be close to $1$.}
\label{table:heu}
\end{table}

\subsection{Training Data Stats}
\label{sec:appendix_traindata}
We share the detailed training data stats in Table~\ref{tab:training_data}.
\begin{table*}[t]
\centering
\fontsize{9pt}{9pt}\selectfont
\setlength{\tabcolsep}{2pt}
\begin{tabular}{lcccccccccccccc}
\toprule
     & Size    & Inst Len & Src Len  & Tar Len  & Len Ratio & Edit Dist & Edit Ratio & Rouge1  \\\midrule
All  & $155676$ & $7.25$   & $30.06$ & $34.47$ & $1.10$    & $20.06$  & $0.30$     & $93.86$ \\ \midrule
Long form & $24384$ & $6.85$   & $118.86$ & $141.09$ & $1.20$    & $115.44$  & $0.97$     & $60.95$ \\
short form  & $131292$  & $7.33$   & $13.60$ & $14.69$ & $1.08$   & $2.35$  & $0.18$ & $99.97$ \\
\bottomrule
\end{tabular}
\caption{Instruction Tuning Training Data Statistics.}
\label{tab:training_data}
\end{table*}

\subsection{Few-shot Prompts for Pre-trained LLM Inference}
\label{subsection:few_shot_prompts_inference}

The few-shot prompt in Table \ref{tab:few_shot_inference} is used to facilitate the pre-trained LLM inference for text rewriting tasks. This is only used for evaluation of our baseline Pre-trained LLMs.

\begin{table*}
\centering
\fontsize{9pt}{9pt}\selectfont
\setlength{\tabcolsep}{1.2pt}
\begin{tabular}{p{0.95\linewidth}}
\toprule
\hspace{0.5cm}Few-shot Prompt\\ 
\midrule

\hspace{0.5cm}\begin{tabular}[c]{@{}p{0.95\linewidth}@{}}Q: make the text more formal: Just ninety-seven more words and I’m done. Why am I misspelling “ninety?” It seems like I shouldn’t misspell it. I should get it right at least once in this paper. There I did it! That gets me within sixty words of being done. It’s funny how I am panicking about my spelling of “panicking.” Wow, that was really not a great sentence. This whole thing has been a strange journey of enlightenment and glee. Yes, I said “glee.” Sixteen more words, and I’ve done it! Keep going, Mitch! Wait, am I not supposed to use my name? Or am I? This is crazy. \\
A: I have to write an additional 97 words before my task is complete. I need to inquire as to why I have misspelled “ninety.” It is a word that I should have the ability to spell correctly. This paper should feature the word's correct spelling at least once. It is amusing that I am rendered anxious due to my inability to spell “anxious” correctly. Surprisingly, the sentence that I recently produced was not high caliber. This task has led me down an odd path of enlightenment and glee. I assure you I said ”glee.” I need to write 16 more words until my task is complete. Maintain your focus, Mitch. Can I state my name? This is a deranged undertaking. \\
Q: make the text more concise: You will also want to get all your legal documents in order. This can range from health and safety certificates to contracts from all employees to contractor’s insurance. For a restaurant, you will likely need a large insurance coverage to allow for any claims that could arise from either staff or customers. \\
A: Get your legal documents in order, from health and safety certificates to contracts with employees to contractor’s insurance. For a restaurant, you will need large insurance coverage to allow for any claims from staff or customers. \\
Q: Please elaborate the following sentence: Next, you should have a menu, or an idea of the cuisine you want to serve at the establishment. You can sit down with some professional contracted chefs to design a menu or you can hire a chef and work on a menu together with the person who will run your kitchen. You will also want to interview people for other major roles within the restaurant, like waiters, sous-chefs, porters, bartenders and hosts. A good manager is key to running the establishment, so you want to make sure you have done your homework and found a suitable candidate that can successfully start your restaurant and maintain it at the level that you desire. \\
A: It would be best to have a menu or a framework for the cuisine you want to serve at your restaurant. You can meet with chefs contracted to create a menu, or you can hire a chef and develop a menu together with the person overseeing your kitchen. It would be best if you were inclined to interview people for other primary roles in your establishment. These roles include servers, sous chefs, bartenders, and hosts. A quality manager is vital to running a restaurant. Ensure you have researched and discovered a qualified candidate who can successfully open your restaurant and maintain it at the desired level. \\ 
Q: Rewrite, reword and reorganize: The issue we need to be aware of is that we can get frustrated trying to find that passion. Though they may work in areas that seem to fit their desires, some think they should wait for the perfect spot before they engage. The idea of “doing something you don’t completely love” is repulsive and a waste of time. \\
A: The issue we need to be cognizant of is that we can get exasperated trying to find that passion. Though they may work in areas that seem to fit their desires, some think they should wait for the perfect spot before they engage. The notion of “doing something you don’t completely love” is revolting and a not worth of one’s time. \\
Q: Please proofread the following sentence: lan om using itjust tobcovwr our car in the winter. \\
A: I plan on using it just to cover our car in the winter. \\
Q: 
\end{tabular}\\

\bottomrule

\end{tabular}
\caption{
The few-shot CoT prompts for pre-trained LLM inference.	}
\label{tab:few_shot_inference}
\end{table*}



\subsection{Few-shot CoT Prompts for LLM Critique}
\label{sec:few_shot_cot_prompts}

The few-shot CoT prompts are used in order to leverage the off-shelf LLMs capability to judge the quality of rewrite. The prompt sample is in Table \ref{tab:critique_prompt}.

\begin{table*}
\centering
\fontsize{9pt}{9pt}\selectfont
\setlength{\tabcolsep}{1.2pt}
\begin{tabular}{p{0.95\linewidth}}
\toprule
\hspace{0.5cm}CoT Prompts\\ 
\midrule

\hspace{0.5cm}\begin{tabular}[c]{@{}p{0.9\linewidth}@{}}Judge whether the \#Response rephrases \#Context and complete the rewriting task in \#Comment. Choose among two choices: GOOD, BAD.\\ \\
\#Comment: Make the text more formal. \\
\#Context: Do we want to go to movie now? This one should be pretty good.\\
\#Response: Want to go to movie? It should be a great one.\\  
\#Choose (GOOD) or (BAD): BAD\\\#Explanation: Response is not more formal than Context.\\ \\ 
\#Comment: Simplify the text. \\
\#Context: Ric Flair had a match against Mitch of the Spirit Squad. All five members of the Spirit Squad were present, so Flair brought out Rowdy Roddy Piper, Money Inc., and Arn Anderson as his backup. Flair's allies kept the Squad in check, enabling Flair to win the match.\\
\#Response: Ric Flair defeated Mitch of the Spirit Squad with help from Rowdy Roddy Piper, Money Inc., and Arn Anderson.\\
\#Choose (GOOD) or (BAD): GOOD\\
\#Explanation: Response is shorter than Context Response preserves overall meaning.\\ \\
\#Comment: Elaborate the following text. \\
\#Context: Iuter X Vanguard collaboration T-shirt by Giorgio Di Salvo. Octopus print. All Iuter apparel is Made in Italy. \\
\#Response: This T-shirt is part of the collaboration between Iuter and Vanguard. It is designed by Giorgio Di Salvo and features an octopus print. All Iuter apparel is Made in Italy. \\
\#Choose (GOOD) or (BAD): GOOD \\
\#Explanation: Response rephrases and elaborates the context with preserved meaning.\\ \\
\#Comment: Paraphrase the source text. \\
\#Context: He likes the dogs a lot, according to his parents. \\
\#Response: He is fond of the dogs. \\
\#Choose (GOOD) or (BAD): BAD \\
\#Explanation: Response did not preserve all the meaning of Context. The fact "according to his parents" is missing in Response.\\ \\
\#Comment: Fix the grammar and spelling error if there is any. \\
\#Context: Native is very fortunate. \\
\#Response: Native people are very fortunate. \\
\#Choose (GOOD) or (BAD): GOOD \\
\#Explanation: Response fix the grammar errors in the Context.\\ \\

\#Comment: \{comment\} \\
\#Context: \{input\}\\ 
\#Response: \{output\_best\}\\ 
\#Choose (GOOD) or (BAD):
\end{tabular}\\

\bottomrule

\end{tabular}
\caption{
The few-shot CoT prompt samples for LLM critique. ``GOOD'' indicates the response is following the instruction of the comment to rewrite the source (context).}
\label{tab:critique_prompt}
\end{table*}

\end{document}